\documentclass{article}
\usepackage{spconf,amsmath,graphicx,paralist}
\usepackage{multirow,url}
\usepackage[caption=false]{subfig}

\title{ADAPTING PRETRAINED TRANSFORMER TO LATTICES\\ FOR SPOKEN LANGUAGE UNDERSTANDING}%
\name{Chao-Wei Huang and Yun-Nung Chen}
\address{National Taiwan University, Taipei, Taiwan\\
\texttt{f07922069@csie.ntu.edu.tw \quad y.v.chen@ieee.org}}

\begin{document}
%
\maketitle
\begin{abstract}

Lattices are compact representations that encode multiple hypotheses, such as speech recognition results or different word segmentations.
It is shown that encoding lattices as opposed to 1-best results generated by automatic speech recognizer (ASR) boosts the performance of spoken language understanding (SLU).
Recently, pre-trained language models with the transformer architecture have achieved the state-of-the-art results on natural language understanding, but their ability of encoding lattices has not been explored.
Therefore, this paper aims at adapting pre-trained transformers to lattice inputs in order to perform understanding tasks specifically for spoken language.
Our experiments on the benchmark ATIS dataset show that fine-tuning pre-trained transformers with lattice inputs yields clear improvement over fine-tuning with 1-best results. Further evaluation demonstrates the effectiveness of our methods under different acoustic conditions\footnote{The code is available at \url{https://github.com/MiuLab/Lattice-SLU}}.

\end{abstract}
\begin{keywords}
Transformer, lattice, spoken language understanding (SLU)
\end{keywords}
\section{Introduction}
\label{sec:intro}

Spoken language understanding (SLU) aims at parsing spoken utterances into corresponding structured semantic concepts.
It plays an important role in spoken dialogue systems, because the whole system may easily fail with the incorrect SLU results.
Typically, SLU includes intent detection and slot prediction.
For example, a movie-related utterance ``{\em find comedies by James Cameron}'' has an intent \textsf{find\_movie} and two slot-value pairs \textsf{(genre, comedy)} and \textsf{(director, James Cameron)}.

An SLU component is usually implemented in a pipeline manner, where spoken utterances are first transcribed by an automatic speech recognizer (ASR), then the transcripts are parsed by a natural language understanding (NLU) system.
One drawback of this approach is that ASR systems may introduce recognition errors, so the following NLU may not be able to capture the correct semantic meaning given the transcribed results.
In addition, the incorrect understanding results from ASR errors may be propagated into later stages of the dialogue system, resulting in undesired responses.

To mitigate this problem, previous work tried to design tighter integration of ASR and NLU systems beyond 1-best results.
The prior work proposed to utilize word confusion networks (WCNs) as input to NLU systems to preserve information in possible hypotheses~\cite{tur2002improving,hakkani2006beyond,henderson2012discriminative,Tr2013SemanticPU}. 
Yaman et al. leveraged n-best lists with similar spirit~\cite{yaman2008integrative}. 
With the recent advance in deep learning methods for SLU \cite{yao2014spoken,guo2014joint,mesnil2014using,goo-etal-2018-slot}, several solutions were proposed to tackle ASR errors.
Masumura et al. examined spoken utterance classification using WCNs with neural networks~\cite{masumura2018neural}.
Inspired by earlier work that extended recurrent neural networks (RNNs) to tree structures~\cite{socher2013recursive,zhu2015long,tai2015improved}, Ladhak et al. proposed a generalized RNN, LatticeRNN, that can process word lattices and achieve better performance for SLU~\cite{ladhak2016latticernn}.
However, due to the inherently sequential nature of RNNs, the training and inference speed of LatticeRNN is dramatically slower than traditional RNNs.

Recently, language models trained with large generic corpora have shown their ability to transfer knowledge from language modeling to various classification tasks either by providing contextualized features~\cite{peters2018deep} or by fine-tuning pre-trained weights on downstream tasks~\cite{Radford2018ImprovingLU,devlin2018bert}.
Fine-tuning a pre-trained transformer~\cite{vaswani2017attention} language model with a small amount of parameters yields state-of-the-art results on a variety of language understanding tasks, such as sentiment classification, textual entailment, and question answering.

Although pre-trained transformers have demonstrated their effectiveness of classifying sequential inputs, it is not clear whether they can encode uncertain inputs in the lattice structure effectively. 
In this work, we aim at exploring the ability of pre-trained transformers to encode and classify lattice inputs. 
Specifically, our main contributions are 3-fold:
\begin{compactitem}
    \item This paper is the first attempt that extends the pre-trained transformer models to take lattice inputs for the spoken language understanding task.
    \item This paper conducts experiments on the benchmark ATIS dataset \cite{hemphill1990atis,dahl1994expanding,tur2010left} and demonstrates clear improvement over baselines using 1-best transcripts.
    \item This paper examines the effectiveness of the proposed method under various acoustic conditions and shows that our method yields consistent improvement.
\end{compactitem}

\section{Language Model Pre-Training}
Language model pre-training has achieved a great success among language understanding tasks with different model architectures.
Because training language models requires a large amount of text data, and it is relatively difficult to acquire a lot of lattices, this work focuses on first pre-training language models with the transformer architecture \cite{vaswani2017attention} and then adapts the model to support lattice inputs.
The pre-training from natural language texts is described below.

\begin{figure}[t!]
    \centering
    \includegraphics[width=0.6\linewidth]{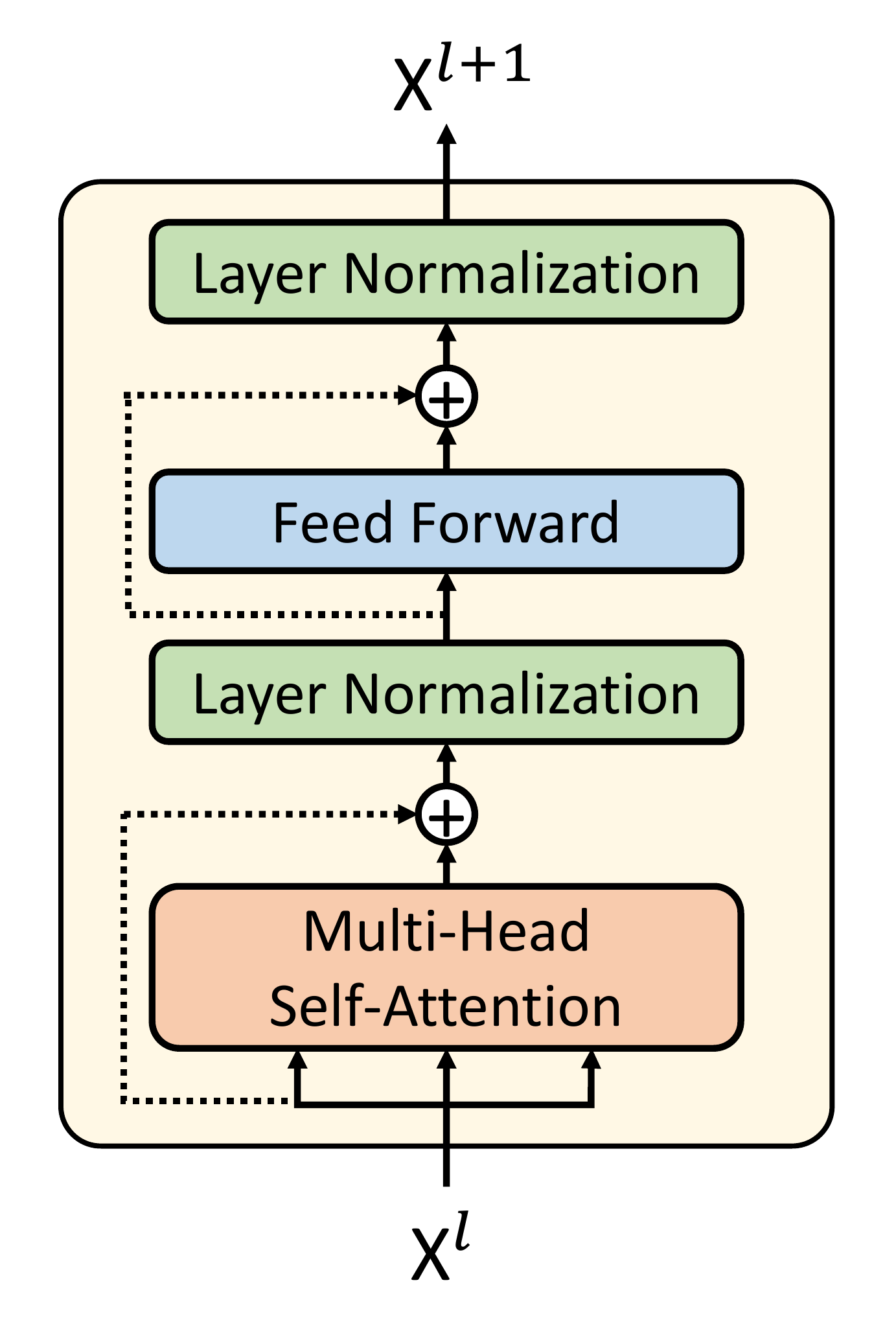}
    \vspace{-3mm}
    \caption{A transformer encoder block}
    \label{fig:transformer}
    \vspace{-2mm}
\end{figure}

\subsection{Transformer Encoder}
We first introduce the transformer encoder model \cite{vaswani2017attention}, which is the backbone model of our method.
The transformer encoder is a stack of $N$ transformer encoder blocks. The $l$-th block takes a sequence of hidden representations $X^l = \{X^l_1, \cdots , X^l_n\}$ as the input and outputs an encoded sequence $X^{l+1} = \{X^{l+1}_1, \cdots , X^{l+1}_n\}$.
A transformer encoder block consists of a multi-head self-attention layer and a position-wise fully connected feed-forward layer. A residual connection \cite{he2016deep} is employed around each of the two layers followed by layer normalization \cite{ba2016layer}.
An illustration of a transformer encoder block is presented in Figure \ref{fig:transformer}.
The detailed components are described as follows.

\subsubsection{Positional Encoding}
Because the transformer model relies on a self-attention mechanism with no recurrence, the model is unaware of the sequential order of inputs.
To provide the model with positional information, positional encodings are applied to the input token embeddings
\begin{equation}
    X^1_i = \text{embed}_{\text{token}}[w_i] + \text{embed}_{\text{pos}}[i],
\end{equation}
where $w_i$ denotes the $i$-th input token, $\text{embed}_{\text{token}}$ and $\text{embed}_{\text{pos}}$ denote a learned token embedding matrix and a learned positional embedding matrix respectively.

\subsubsection{Multi-Head Self-attention}
An attention function can be described as mapping a query to an output with a set of key-value pairs. The output is a weighted sum of values.
We denote queries, keys and values as $Q$, $K$ and $V$, respectively.
Following the original implementation \cite{vaswani2017attention}, a scaled dot-product attention is employed as the attention function. 
Hence, the output can be calculated as
\begin{equation}
    \text{Attention}(Q, K, V) = \text{softmax}(\frac{QK^T}{\sqrt{d_k}})V,
\end{equation}
where $d_k$ denotes the dimension of key vectors.

The idea of multi-head attention is to compute multiple independent attention heads in parallel, and then concatenate the results and project again. 
The multi-head self-attention in the $l$-th block can be calculated as
\begin{align}
    & \text{MultiHead}(X^l) = \text{Concat}(\text{head}_1, \cdots, \text{head}_h)W^O, \\
    & \text{head}_i = \text{Attention}(X^lW^Q_i, X^lW^K_i, X^lW^V_i),
\end{align}
where $X^l$ denotes the input sequence of the $l$-th block, $h$ denotes the number of heads, $W^Q_i$, $W^K_i$, $W^V_i$ and $W^O$ are parameter matrices.

\subsubsection{Position-Wise Feed-Forward Layer}
The second sublayer in a block is a position-wise feed-forward layer, which is applied to each position separately and independently.
The output of this layer can be calculated as
\begin{equation}
    \text{FFN}(x) = \max(0, x\cdot W_1 + b_1)W_2 + b_2,
\end{equation}
where $W_1$ and $W_2$ are parameter matrices, $b_1$ and $b_2$ are parameter biases.

\subsubsection{Residual Connection and Layer Normalization}
As shown in Figure~\ref{fig:transformer}, the residual connection is added around the two sublayers followed by layer normalization. The output of the $l$-th block can be calculated as
\begin{align}
    H^{l} & = \text{LayerNorm}(\text{MultiHead}(X^l) + X^l), \\
    X^{l+1}  & = \text{LayerNorm}(\text{FFN}(H^l) + H^l).
\end{align}

\begin{figure}
    \centering
    \subfloat[The pre-training stage.]{\includegraphics[width=0.75\linewidth]{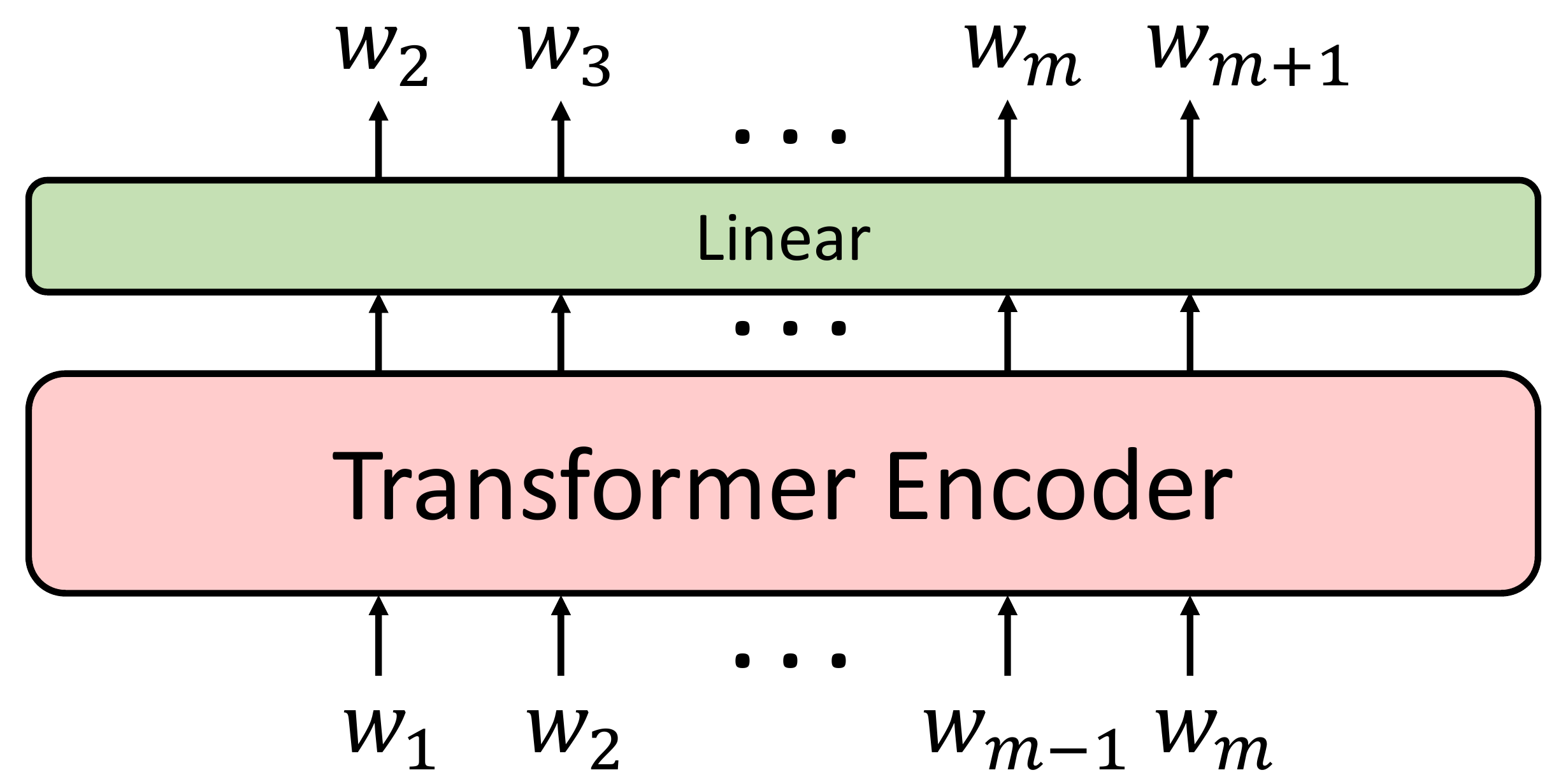}} \\
    \subfloat[The fine-tuning stage.]{\includegraphics[width=0.75\linewidth]{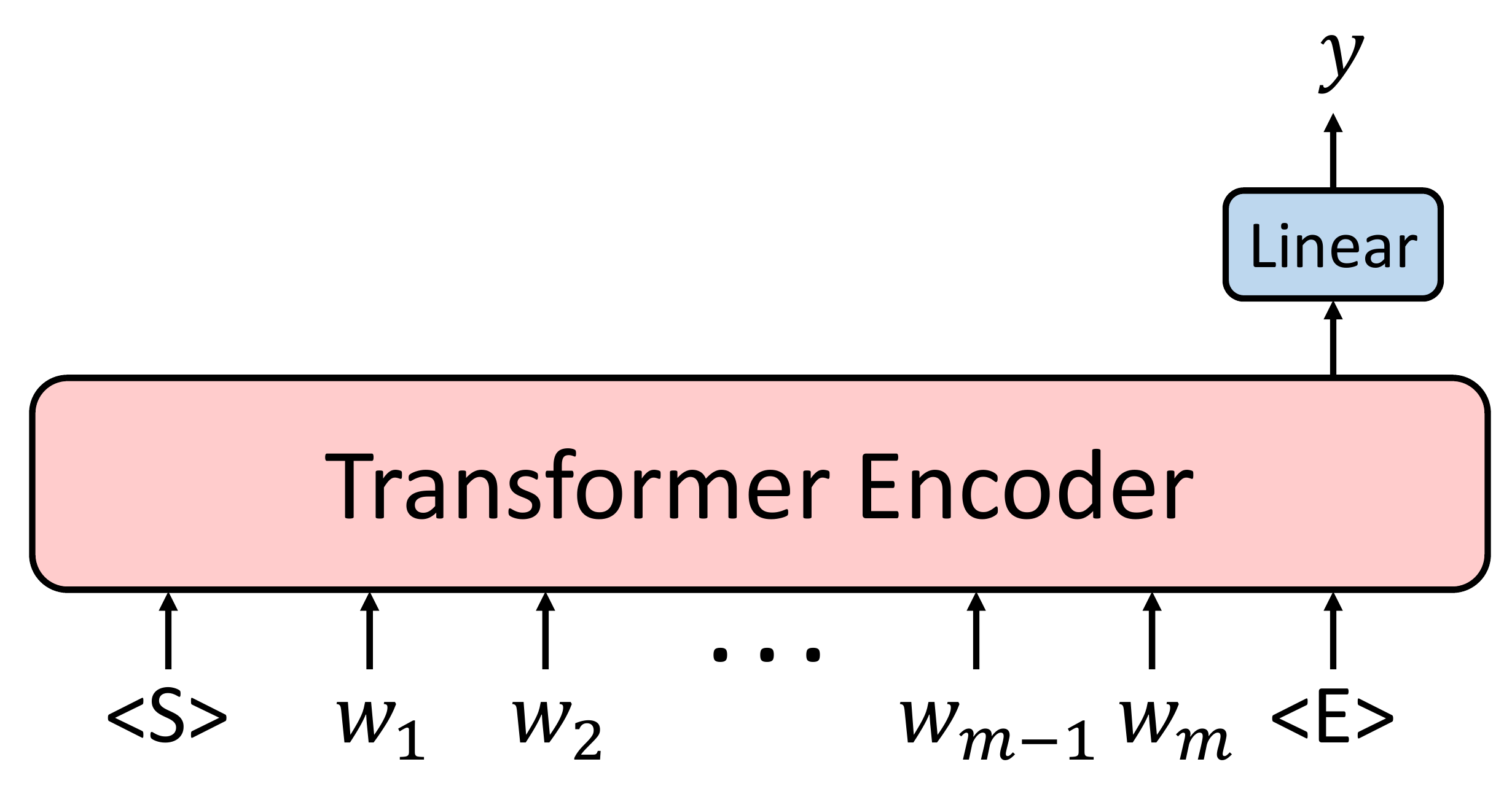}}
    \caption{Illustration of the two-stage method in the GPT model. $\texttt{<S>}$ and $\texttt{<E>}$ are special tokens introduced when fine-tuning.}
    \label{fig:GPT}
\end{figure}

\subsection{Generative Pre-Training Model (GPT)}
The generative pre-training (GPT) via a language model objective is shown to be effective for learning representations that capture syntactic and semantic information without supervision \cite{peters2018deep,Radford2018ImprovingLU,devlin2018bert}.
The GPT model proposed by Radford~\cite{Radford2018ImprovingLU} employs the transformer encoder with 12 encoder blocks. 
It is pre-trained on a large generic corpus that covers a wide range of topics.
The training objective is to minimize the negative log-likelihood:
\begin{equation}
    \mathcal{L} = \sum_{t=1}^{T}{-\log P(w_t \mid w_{<t}, \theta)},
\end{equation}
where $w_t$ denotes the $t$-th word in the sentence, $w_{<t}$ denotes all words prior to $w_t$, and $\theta$ is parameters of the transformer model.

To avoid seeing the future contexts, a masked self-attention is applied to the encoding process. 
In the masked self-attention, the attention function is modified into
\begin{equation}
    \text{Attention}(Q, K, V) = \text{softmax}(\frac{QK^T}{\sqrt{d_k}} + M)V,
\end{equation}
where $M$ is a matrix representing masks.
$M_{ij} = -\infty$ indicates that the $j$-th token has no contribution to the output of the $i$-th token, so it is essentially ``\emph{masked out}'' when encoding the $i$-th token.
Therefore, by setting $M_{ij} = -\infty$ for all $j > i$, we can calculate all outputs simultaneously without looking at future contexts.

After the model is pre-trained with a language model objective, it can be fine-tuned on downstream tasks with supervised data. Given a sequence of input tokens $w_1, \cdots, w_m$ along with label $y$, the inputs are passed through the pre-trained encoder to obtain the output of the last token $X^{12}_{m}$, which is then fed into a linear layer with parameters $W_y$ to make predictions. Note that the linear layer is added in the fine-tuning stage. 
Figure~\ref{fig:GPT} illustrates this two-stage approach.
By fine-tuning on the target tasks, the GPT model has achieved the state-of-the-art performance on various supervised tasks including natural language inference, question answering, semantic similarity and linguistic acceptability \cite{Radford2018ImprovingLU}.

\section{Adapting Pre-trained Transformer to Lattices}

As mentioned before, pre-training the language models from large lattice data is difficult.
Therefore, our approach focuses on proposing an adaptation method that allows the pre-trained transformer to take lattices as its input.
A simple approach of enabling the pre-trained transformers to consume lattice inputs is to flatten the lattice by its topological order and directly apply the sequential transformer model~\cite{sperber2019selfattentional}.
However, this approach ignores the graph structure entirely, and the resulting sequence may not form a meaningful sentence or may even express the different meaning.
Sperber et al.~\cite{sperber2019selfattentional} introduced \textit{lattice reachability masks} and \textit{lattice positional encoding} to encode lattices with self-attentional models. 
In the following subsections, we define the lattices and describe how we integrate the above techniques into the pre-trained transformer model in detail.

\begin{figure}[t!]
    \centering
    \includegraphics[width=\linewidth]{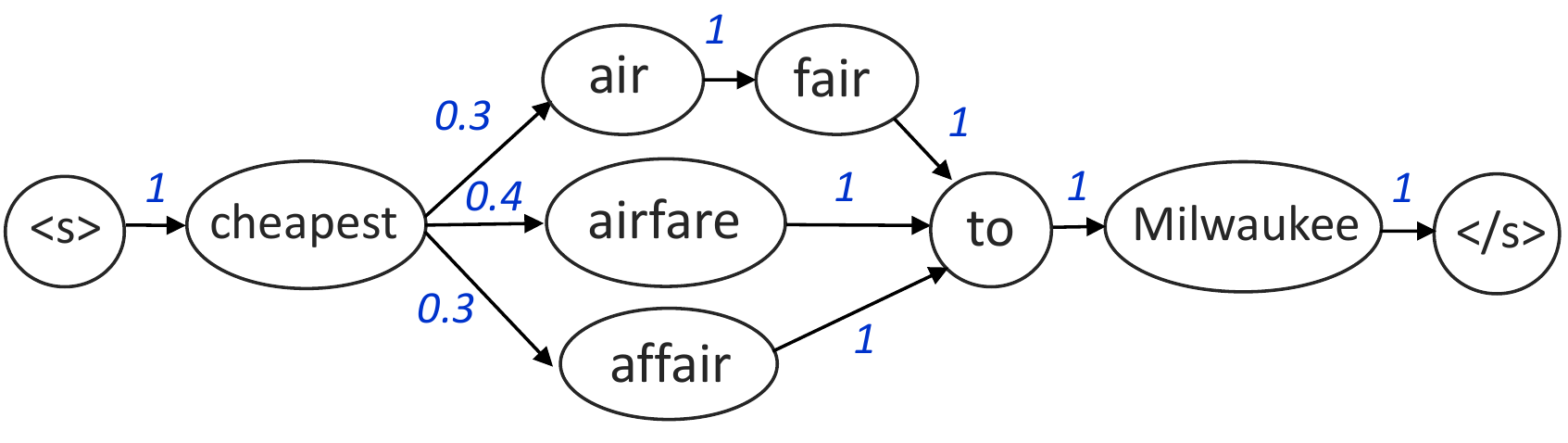}
    \caption{An example lattice representation. Numbers in blue indicate the transition probabilities.}
    \label{fig:lattice}
\end{figure}

\subsection{Lattices}
Lattices represent multiple competing sequences in a compact structure as directed acyclic graphs (DAGs). There is one start node, labeled as \texttt{\textless s\textgreater}, and one end node, labeled as \texttt{\textless /s\textgreater}. Each node is labeled as a token. In our case, lattices are outputs of ASR that store multiple decoded hypotheses with uncertainty, and each path from the start node to the end node indicates a possible hypothesis. An example lattice is illustrated in Figure \ref{fig:lattice}.

More formally, let $G = (V, E)$ be a DAG with nodes $V$ and edges $E$. For a node $v \in V$, we denote the set of its predecessors as $\text{Pre}(v)$. 
For any pair of nodes $v_i$ and $v_j$, \\ $\text{P}(v_j \in \text{Pre}(v_i) \mid v_i)$ represents the conditional probability that a path from start node to end node in $G$ contains $v_j$ as a predecessor of $v_i$, given that $v_i$ appears in the path.

\subsection{Lattice Reachability Masks}
Recall that lattices store multiple hypotheses in a DAG.
If we perform self-attention with respect to all nodes in the lattice, the information from different hypotheses may be mixed in an undesired way, because the model is unaware of the global structure of lattices due to the fact that it is pre-trained with sequential data only.

Inspired by the concept about conditioning from TreeLSTM \cite{tai2015improved} where each node is conditioned on its predecessors, Sperber et al. proposed \textit{lattice reachability masks} to prevent the self-attention mechanism from attending to nodes that are not predecessors of a given query node $v_i$ \cite{sperber2019selfattentional}.
There are two variants of masks:
\begin{itemize}
    \item \textbf{Binary masks} restrict the self-attention mechanism to only attend to the predecessors of the query node $v_i$, other nodes are masked:
    \begin{equation}
        M^{bin}_{ij} = 
        \begin{cases}
            0 & v_j \in \text{Pre}(v_i) \text{ or } v_i = v_j, \\
            -\infty & \text{otherwise.}
        \end{cases}
    \end{equation}
    \item \textbf{Probabilistic masks} generalize the binary masks to a probabilistic form. Binary masks ignore the uncertainties in the lattice, so the model considers each node equally disregarding their confidence scores. Therefore, probabilistic masks are designed to bias the attention toward the nodes with higher confidence:
    \begin{equation}
        M^{prob}_{ij} = 
        \begin{cases}
            \log P (v_j \in \text{Pre}(v_i) \mid v_i)     & v_j \in \text{Pre}(v_i), \\
            0   & v_i = v_j, \\
            -\infty & \text{otherwise.}
        \end{cases}
    \end{equation}
\end{itemize}

In the pre-training stage, a mask $M$ is used to prevent the tokens from attending to future contexts, where $M_{ij} = -\infty$ for all $j > i$. Notice that the masks $M^{bin}$ and $M^{prob}$ described above are strict generalization of $M$, considering that the successors of the query node is always masked.
In order to leverage the information from lattices, this paper proposes to extend the masked self-attention to
\begin{equation}
    \text{Attention}(Q, K, V) = \text{softmax}(\frac{QK^T}{\sqrt{d_k}} + M^{lattice})V,
\end{equation}
where $M^{lattice}$ can be either $M^{bin}$ or $M^{prob}$.
Therefore, the signal about acoustic confidence from lattices provides additional cues for the model, and it may perform better for spoken language tasks due to the lattice information.

\subsection{Lattice Positional Encoding}
Positional encoding is crucial for transformer models, since it is the only source providing token ordering information. Normally, transformers use either learned or fixed embeddings for each position. 
When dealing with lattice inputs, determining the position of each lattice node is not straight-forward.
Sperber et al.~\cite{sperber2019selfattentional} proposed to use the longest-path distance from the start node as a node's position, so the input of the transformer model becomes
\begin{equation}
    X^1_i = \text{embed}_{\text{token}}[w_i] + \text{embed}_{\text{pos}}[\text{ldist}(v_i)],
\end{equation}
where $\text{ldist}(v_i)$ denotes the longest-path distance from the start node to the node $v_i$ and $w_i$ is the label of $v_i$.

\begin{figure}[t!]
\centering
\includegraphics[width=0.9\linewidth]{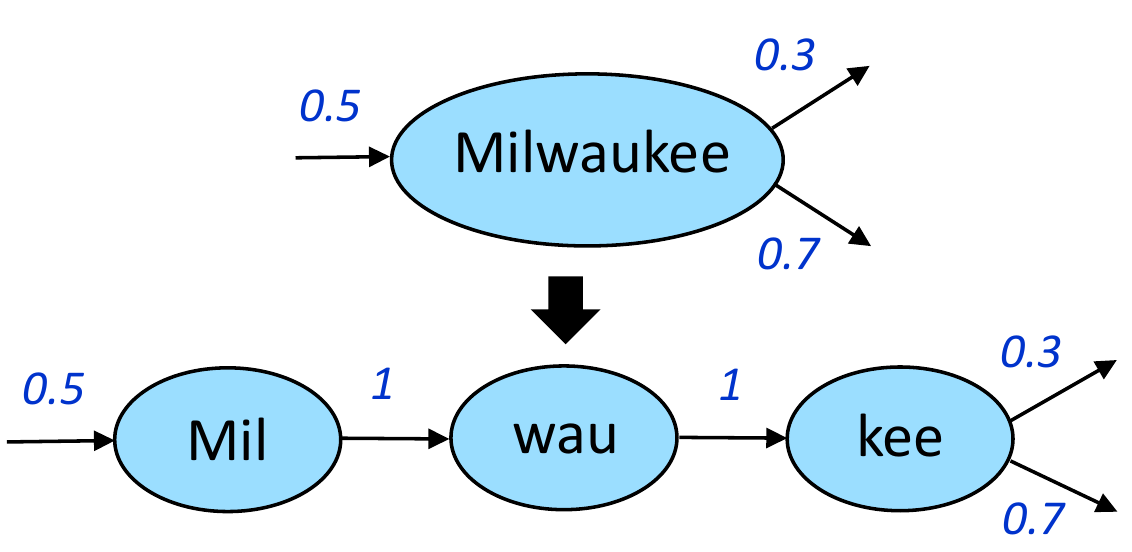}
\caption{The illustration of the node-splitting method, where the original node with the label \texttt{Milwaukee} is split into nodes with tokenized labels.}
\label{fig:lattice-split}
\end{figure}

This method that assigns the position to lattice nodes brings several benefits:
1) positions of each lattice path are strictly monotonically increasing, which complies with the setting in pre-training,
2) unnecessary jump between neighboring nodes are avoided, and
3) positions are no larger than the length of the longest hypothesis, so the sequence lengths do not differ much from that in pre-training.

The pre-trained GPT model uses byte-pair encoding to split words into subword units. To follow this tokenization scheme, we split a node in word lattice generated by ASR according to the tokenization of its label. 
For instance, a node labeled as \textsf{Milwaukee} is split into three nodes that are labeled as \textsf{Mil}, \textsf{wau}, \textsf{kee} respectively. 
An example of this processing method is shown in Figure \ref{fig:lattice-split}. 
After node splitting, we assign positions to the resulting lattice using the \emph{longest-path distance} method.

\begin{table*}[t]
\centering
\caption{F1-scores for multi-label classification and accuracy for utterance-level performance on ATIS (\%).}
\label{tab:result}
\vspace{2mm}
\begin{tabular}{|cl|c|c|c|c|}
\hline
\multicolumn{2}{|c|}{} & \multicolumn{2}{c|}{\bf Intent} & \multicolumn{2}{c|}{\bf Slot} \\ \cline{3-6}
\multicolumn{2}{|c|}{} & \multicolumn{1}{c|}{\bf F-Measure}      & \multicolumn{1}{c|}{\bf Accuracy} & \multicolumn{1}{c|}{\bf F-Measure} & \multicolumn{1}{c|}{\bf Accuracy} \\
                        \hline\hline
\multirow{4}{*}{Condition 1 (WER=15.5\%)} & 1-Best Baseline        & 97.38        & 96.30       & 93.76       & 76.01      \\
                        & Lattice-Linearize      &  97.98    &  96.90  &   94.56       &    79.21        \\
                        
                        & Lattice-Probabilistic   &  98.19        & \bf 97.21       & \bf 94.65       & 79.45      \\
                        & Lattice-Binary & \bf 98.23        & 97.15       & \bf 94.65       & \bf 79.93      \\
                        \hline\hline
\multirow{4}{*}{Condition 2 (WER=26.3\%)} & 1-Best Baseline    & 94.25        & 92.67       & 87.98       & 60.23      \\
                        & Lattice-Linearize      &   94.87        &     93.32        & 88.59        & 62.14           \\
                        
                        & Lattice-Probabilistic   & 95.14        & 93.40        & 89.01       & 62.86      \\
                        & Lattice-Binary & \bf 95.25        & \bf 93.50        & \bf 89.09       & \bf 63.43      \\
                        \hline\hline
\multirow{4}{*}{Condition 3 (WER=38.7\%)} & 1-Best Baseline        & 90.64        & 86.40       & 87.31       & 58.59      \\
                        & Lattice-Linearize      &   91.87      &  88.14  &     88.36        & 59.58           \\
                        & Lattice-Probabilistic   & \bf 92.57        & \bf 89.48       & \bf 88.67       & \bf 61.38      \\
                        & Lattice-Binary & 92.39        & 89.06       & 88.66       & 60.98      \\
                        \hline\hline
                        \multicolumn{2}{|c|}{Reference} & 98.90 & 98.08 & 95.96 & 87.14 \\
                        \hline 
\end{tabular}
\end{table*}

\section{Experiments}
We conduct the experiments on a benchmark SLU task to examine the effectiveness of our method, where we fine-tune the pre-trained transformer model with lattice inputs.

\subsection{Setup}
ATIS (Airline Travel Information Systems) ~\cite{hemphill1990atis,dahl1994expanding,tur2010left} is a widely used dataset for benchmarking SLU research.
The dataset contains audio recordings of people making flight reservations or asking flight information with corresponding manual transcripts.
The training set contains 4,478 utterances and the test set contains 893 utterances.
We treat both intent detection and slot prediction as multi-label classification problems, where slot prediction tries to predict what kind of slots appear in an utterance.
There are 81 slot labels and 18 intents in the training set.

Our ASR is trained on WSJ~\cite{Paul:1992:DWS:1075527.1075614} using the \textit{s5} recipe from Kaldi~\cite{povey2011kaldi}.
We use the ASR system to recognize audio recordings in ATIS training set and extract lattices for fine-tuning. 
To simulate different acoustic conditions, we artificially corrupt the recordings with additive noises from MUSAN corpus \cite{musan2015} and simulated room impulse response \cite{allen1979image,ko2017study}. 
As a result, we have three copies of the dataset:
\begin{compactitem}
    \item \textit{Condition 1}: The original dataset. The word error rate (WER) of the ASR results is 15.55\% in the test set.
    \item \textit{Condition 2}: A mildly corrupted version of the original dataset. The WER is 26.30\% in the test set.
    \item \textit{Condition 3}: A severely corrupted version of the original dataset. The WER is 38.69\% in the test set.
\end{compactitem}

\subsection{Model and Training Details}
The pre-trained weights of GPT from the original paper are adopted~\cite{Radford2018ImprovingLU}.
A linear classifier is placed on top of the outputs of  transformers, and we use the output of the last token for prediction.

When fine-tuning GPT, we set the batch size to 8 and use \texttt{Adam} as the optimizer \cite{kingma2014adam} with learning rate $6.25 \cdot 10^{-5}$. 
A linear warm-up schedule is adopted and the whole model is fine-tuned for 5 epochs.
We train intent detection and slot prediction models separately.

\subsection{Results}
Table~\ref{tab:result} presents the results of intent detection and slot prediction under three acoustic conditions. 
For each number in the table, we perform ten runs with different random seeds and calculate the average after discarding the best and the worst runs.
In each condition, we compare our methods with two baseline systems: \textbf{1-best} that takes 1-best transcripts as the input and \textbf{lattice-linearize} that takes lattices linearized with topological order.

The results show that fine-tuning pre-trained transformers with lattices significantly outperforms 1-best baseline on both tasks, yielding 32.4\%, 15.4\% and 20.6\% relative error reduction under different acoustic conditions.
The two variants of masks perform similarly across acoustic conditions.
These results demonstrate that fine-tuning pre-trained transformers with lattices yields clear and consistent improvement for SLU.

In addition, using linearized lattices also improves the performance over the 1-best baseline. 
This is probably that ATIS is relatively simple, where keywords play an important role for understanding.
Therefore, even though the linearized lattices do not form meaningful sentences, they still provide the model with more possible words and achieve the improved performance.
Nevertheless, the best performance in the experiments comes from our proposed methods.

\begin{table*}[t]
\centering
\caption{A testing sample in the condition 3.}
\label{tab:sample}
\vspace{2mm}
\begin{tabular}{llc}
\hline
 & \bf Input & \bf SLU Result\\
\hline
\textbf{Reference} & how far is new york 's la guardia from downtown & \textit{distance}\\
 \textbf{ASR 1-best} & how for is new york 's la guardia from downtown &  None\\
 \textbf{ASR Lattice} &
\includegraphics[width=0.5\linewidth]{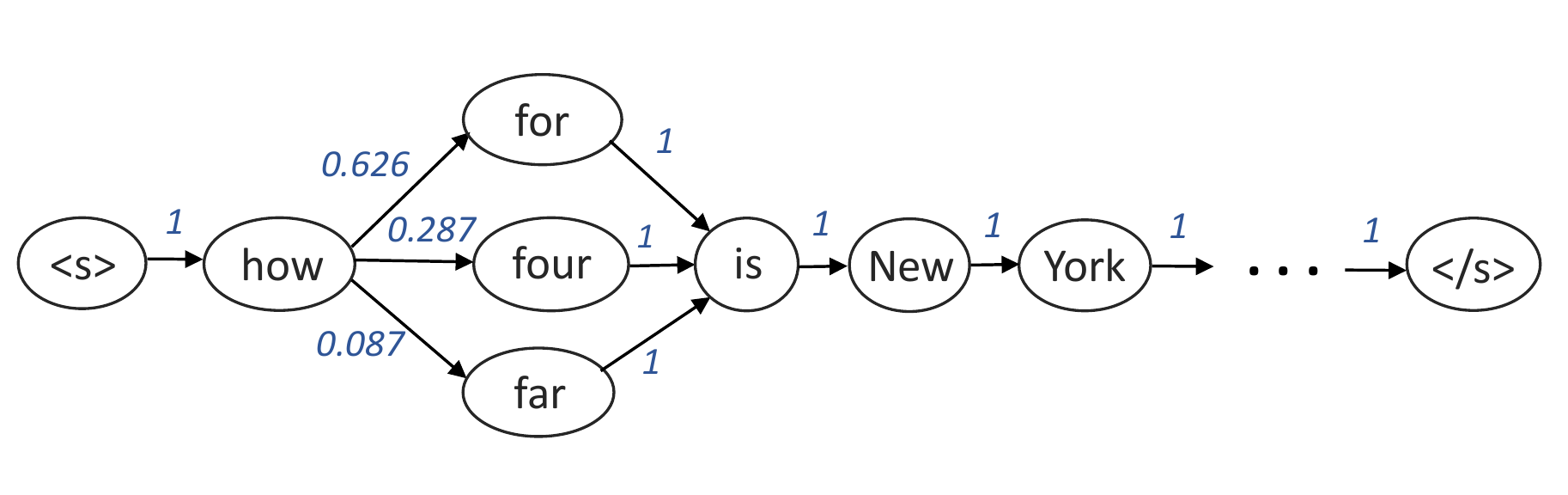} & \it distance\\
\hline
\end{tabular}
\vspace{-2mm}
\end{table*}

\section{Discussion and Analysis}

We investigate the sample efficiency and ASR impact of our model and perform qualitative analysis here.

\begin{figure}
    \centering
    \includegraphics[width=\linewidth]{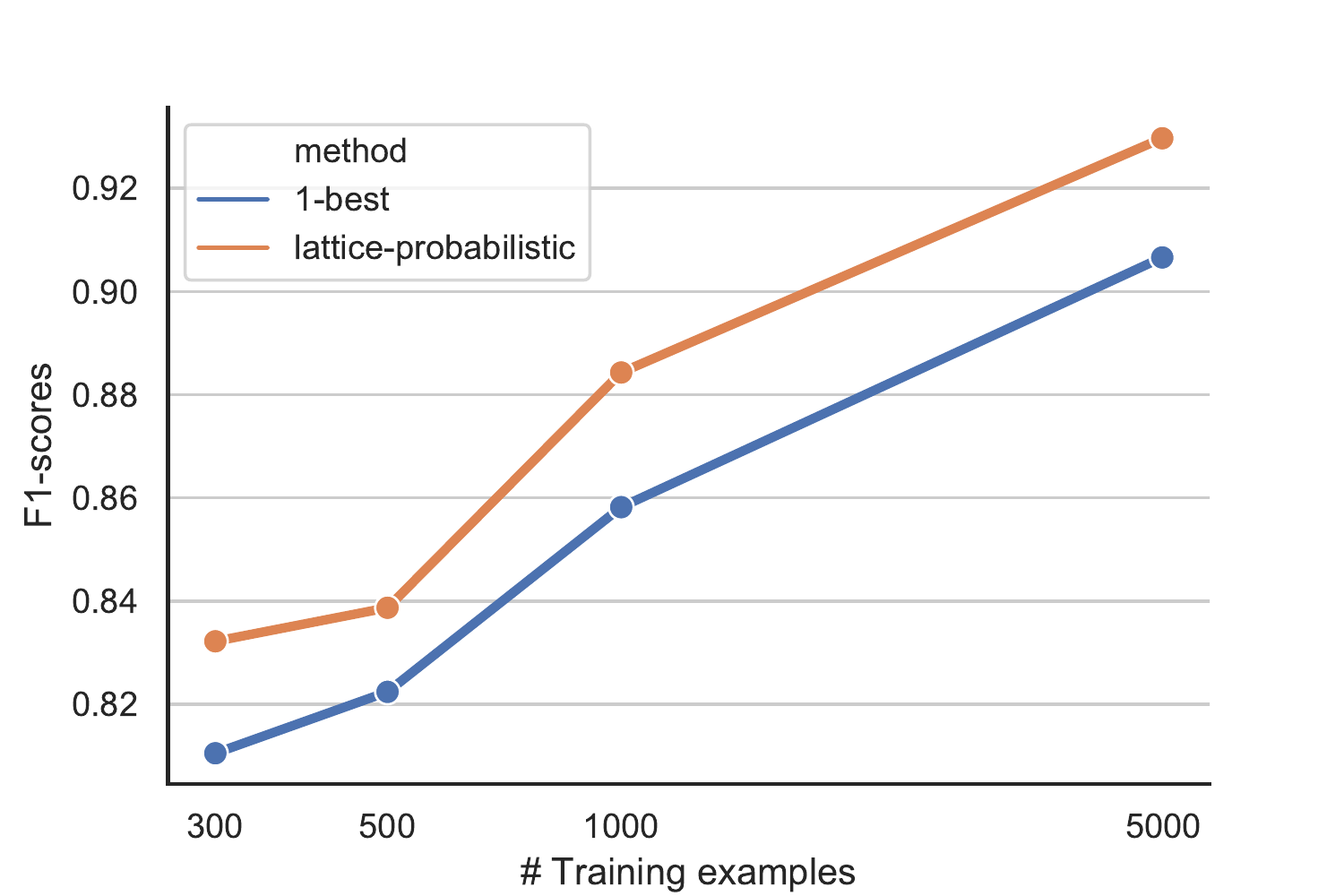}
    \vspace{-2mm}
    \caption{F1 with respect to size of training data.}
    \label{fig:data-eff}
\end{figure}

\subsection{Sample Efficiency}
Although lattices provide richer information than 1-best transcripts so the model can reach better performance, it is possible that the model requires more examples to learn the proper mapping between utterances and labels due to the noisy input lattices.
We conduct experiments with different numbers of training examples to test the sample efficiency of fine-tuning with lattices. We sample $n$ examples randomly from the original training set and use this reduced set to fine-tune our models. All examples are drawn from the condition 3.

Figure~\ref{fig:data-eff} plots the results, where fine-tuning with lattices consistently outperforms the 1-best baseline by 2\% while the number of training examples varies from 300 to 5000, indicating that our method does not require more training examples to achieve such improvement.

\begin{figure}
    \centering
    \includegraphics[width=\linewidth]{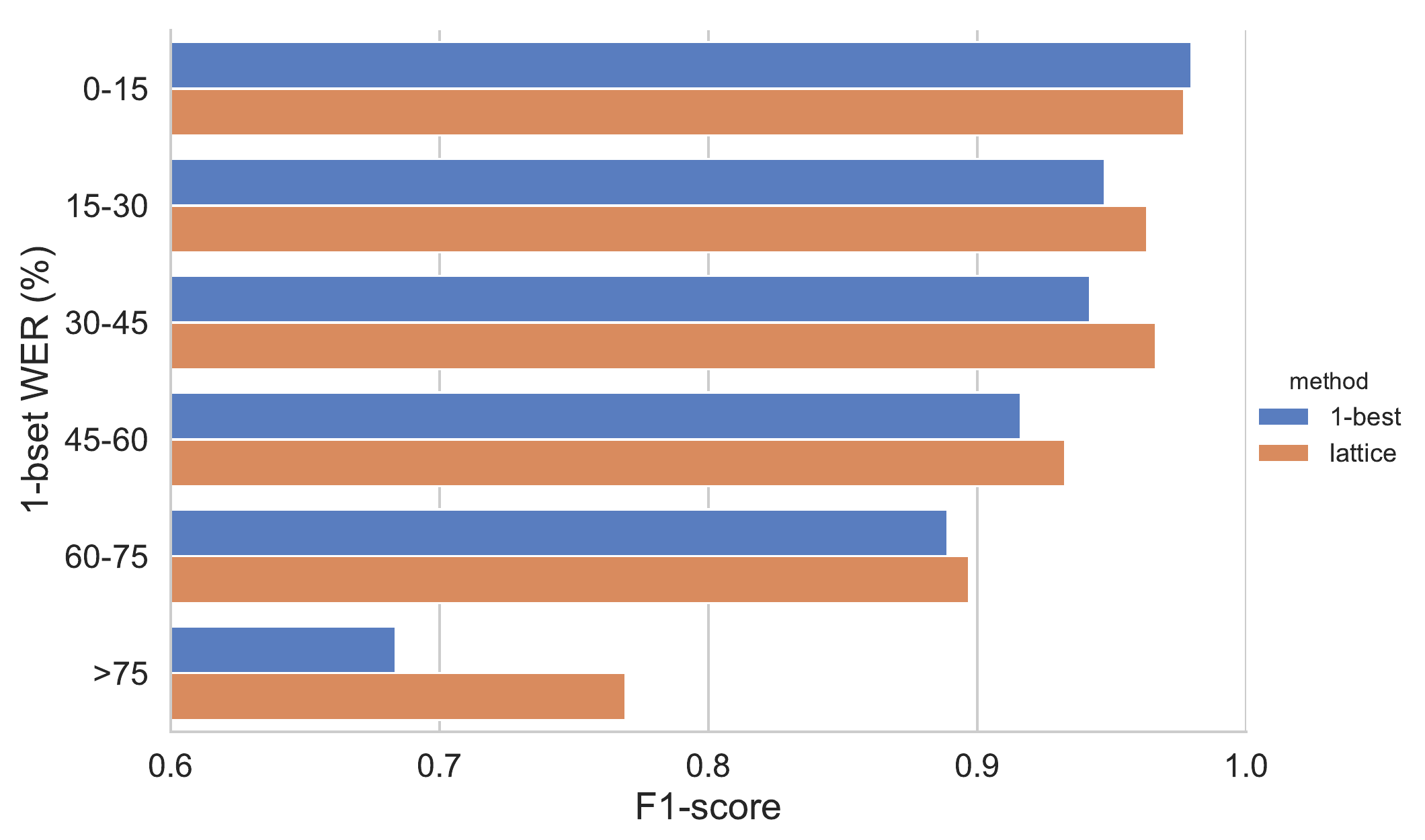}
    \vspace{-3mm}
    \caption{F1 with respect to utterance-level word error rates.}
    \label{fig:wer-sensitivity}
\end{figure}

\subsection{Impact of Transcription Quality}
We analyze the SLU performance with respect to WER of ASR.
We group the utterances according to their utterance-level WER and present the performance of each group in Figure~\ref{fig:wer-sensitivity}.
Aligning with our intuition, using 1-best results achieves slightly better performance when lower WER ($<$15\%).
In all other cases, using lattices as inputs outperforms using 1-best, demonstrating that our model is especially suitable for the scenarios with poor ASR results.

\subsection{Qualitative Analysis}
In order to further analyze how lattices help SLU, we sample an example from the test set of the condition 3 shown in Table~\ref{tab:sample}.
In this example, the ASR misrecognizes \textit{far} as \textit{for}, so the SLU is unable to understand the intention behind this utterance. 
It is clear that the correct word, \textit{far}, is in the lattice, so SLU is able to correctly understand that the user is asking about the distance, even though \textit{far} has the lowest probability among alternatives.
It justifies the effectiveness of our model.

\section{Conclusion}
This paper extends the pre-trained transformer to lattice inputs in order to perform understanding on lattices generated by ASR systems.
We leverage lattice reachability masks and lattice positional encoding into the pre-trained transformer model, enabling it to consume lattice inputs during fine-tuning.
The experiments on benchmark SLU data demonstrate the effectiveness of our methods under various acoustic conditions\footnote{This work was financially supported from the Young Scholar Fellowship Program by MOST in Taiwan, under Grant 108-2636-E002-003.}.


\bibliographystyle{IEEEbib}
\bibliography{strings,refs}

\end{document}